\documentclass{article}
\usepackage{spconf,amsmath,epsfig}

\title{Bayesian approach for near-duplicate image detection}
\name{Lucas Moutinho Bueno, Eduardo Valle, Ricardo Torres \thanks{Thanks to FAPESP for funding.}}
\address{University of Campinas, Institute of Computing, Brazil}
\begin{document}
%\ninept
%
\maketitle
\begin{abstract}

In this paper we propose a bayesian approach for near-duplicate image detection, and investigate how different probabilistic models affect the performance obtained. The task of identifying an image whose metadata are missing is often demanded for a myriad of applications: metadata retrieval in cultural institutions, detection of copyright violations, investigation of latent cross-links in archives and libraries, duplicate elimination in storage management, etc. The majority of current solutions are based either on voting algorithms, which are very precise, but expensive; either on the use of visual dictionaries, which are efficient, but less precise. Our approach, uses local descriptors in a novel way, which by a careful application of decision theory, allows a very fine control of the compromise between precision and efficiency. In addition, the method attains a great compromise between those two axes, with more than 99\% accuracy with less than 10 database operations.

\end{abstract}
%
%\section{Introduction}
%\label{sec:intro}

\section{INTRODUCTION}

Identifying an image whose part of data is unknown (near-duplicate image detection) is demanded for many applications. Usually, very large image datasets are involved, ranging from tens of thousands to hundreds of millions images. What makes the task of image identification complex is that the query image, which one is trying to identify, has usually has suffered transformations from the reference image in the dataset. Those transformations include croppings, changes of scale, rotations, non-affine geometric transformations, photometric and colorimetric changes, compression, occlusions and other assorted transforms, like dithering and fancy artistic effects.\\
\indent The most reliable solutions to image identification today employ local features in a way or another. Local features present a remarkable robustness to geometric, photometric and colorimetric transforms, and, because of that, and because of their sheer density over a single image, they provide a very powerful scheme to match  the same object or scene among different images. The majority of current solutions are based either on voting algorithms, which are very precise, but expensive; either on the use of visual dictionaries, which are efficient, but less precise.\\
\indent It is remarkable that both solutions make a very indirect use of the distance between features in the feature space: vote algorithms usually take into consideration the nearest matches between query features and database features, but then do not take into account the actual distance between the matched features (except for establishing a contrast threshold, see \cite{SIFT}). Dictionary approaches often use the distance to a set of prototypes to establish the image description, but the actual distance value is lost after the description is encoded.\\
\indent The approach we proposed is at once very simple and different, in which we propose to give a direct interpretation to the distance between the query features and the database features, in terms of how sure we are to images form a true correspondence. To establish this interpretation, we use the elegant framework of Bayesian decision theory. This allows to obtain a very efficient scheme, with very database interrogations (contrarily to raw vote-based systems) and a very fine control of the compromise between precision and efficiency (contrarily to dictionary-based systems). Our experiments show how promising the scheme is, with more than 99\% accuracy with less than 10 database operations.\\
\indent Another important contribution of this work is the investigation of the impact of different probabilistic models on the performance obtained. We show that, contrarily to the Gaussian model usually employed, SIFT descriptors follow a Chi distribution with an excellent fitting. The experiments clearly demonstrate how passing from the more general model (Gaussian) to the more specific (Chi) improves the accuracy of the system. We believe that this observation might impact other applications of SIFT features beyond nearest-duplicate detection.

\section{RELATED WORK}

Local image descriptors describe visual features around interest points such as blobs and corners and represent then by a feature vector. Hundreds to a few thousand of interest points can be found on a single image.

Local features are especially effective for applications that do not need the generalization power of category search. Those applications, aimed at target matching, may involve the recognition of specific objects, scenes or images.

In those applications, a match between local features are highly indicative of a match between objects or scenes. Usually, those matches are obtained simply by taking the local feature in the dataset which has minimum distance to the query feature (an operation called nearest neighbor query).  To avoid false positives, other criteria may be imposed, like requiring the matches to be geometrically consistent or using a criterion of contrast to warrant that the match is distinctive \cite{SIFT}.

Descriptors are expected to be invariant to image transformations (geometric or radiometric) and highly distinctive. Many good quality local image descriptors have been proposed on literature on the past few years. Among then we can cite: SIFT \cite{SIFT}, PCA-SIFT \cite{PCASIFT}, GLOH \cite{DesComp}, SURF \cite{SURF}. Those ones are gradient based descriptors and have shown more robust and distinctive for target matching applications then spectral based descriptors \cite{DesComp}.

Since all of those descriptors above cited presented good results according to its references, we chose to use SIFT on our experiments because it’s the most known and referenced local descriptor. The standard SIFT detector (Difference of Gaussians - DoG) were used to find interest points.

\subsection{Near-Duplicate Detection}

Near-duplicate detection is a intensely studied research topic, with a huge literature \cite{NDIR1,NDIR2,NDIR3,NDIR4,NDIR5}, which consists on finding an original image on a large database from a transformed query image. We based our work on previous voting-based systems presented on \cite{NDIR4} and refined on \cite{NDIR2}. That system is based on taking each and every feature of the query image and matching it with its nearest neighbor on the dataset, retrieving the image with most matches. See Figure 1.

Vote-based Systems take much time to retrieve an image for a single query because of the huge number of distances computed between feature vectors. Specially-designed indexes and fast approximate matching were proposed to alleviate the burden of matching the datasets \cite{NDIR2,NDIR3}, but still, hundreds of query operations must be performed. Another popular solution, is to compact the representation of multiple features into a single "bag of visual features" representation, that can then be queried \cite{NDIR5}. That latter solution is very fast, but one loses on precision.

\begin{figure}[htb]

\begin{minipage}[b]{1.0\linewidth}
  \centering
  \centerline{\epsfig{figure=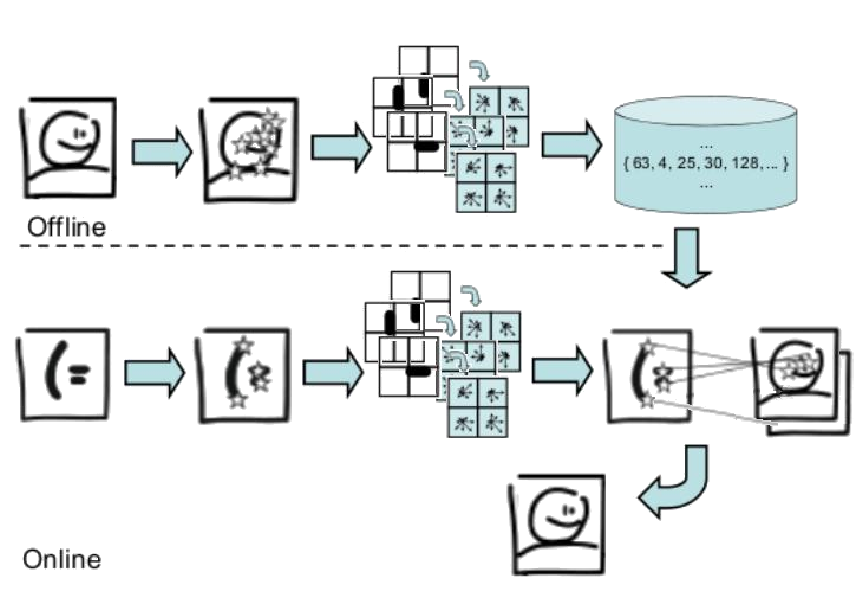,width=8cm}}

\end{minipage}

\caption{The voting system for near duplicate detection}
\end{figure}

\section{PROPOSED SCHEME}

Here we propose to use Bayesian decision theory, to take into account the observed distances in the matches. This allows us to reach an appropriate decision after matching less than a dozen feature vectors from the query image.

In this preliminary work, we match the features exactly, but the decision model is fully compatible with approximate matching, and we are currently working on incorporating the accelerated index in the scheme.

\subsection{Training}

 We first extract the feature vectors on a set of query images and perform a nearest neighbors search from their feature vectors and a database containing the original images (near-duplicates, before transformation) and a confounding set of unrelated images (noise), used to try to perturb the match. After the match is done, we separate them in two populations: those that correspond to features between "correct" images (transformed images and their originals) and those between "incorrect" (unrelated) images. We compute, then, for each of those two populations, an histogram of the distances (in the feature space) between the query and the target feature.

Figure 2 shows both histograms. As we can see the histogram of correct match distances (Figures 2a and 2b) has a curve much closer to zero then the incorrect one (Figures 2c and 2d), which means that the distance between feature vectors for the first case has highly frequently lower values than for the second case, as is expected. 

\begin{figure}[htbp]

\begin{minipage}[h]{1 \linewidth}
  \centering
  \centerline{\epsfig{figure=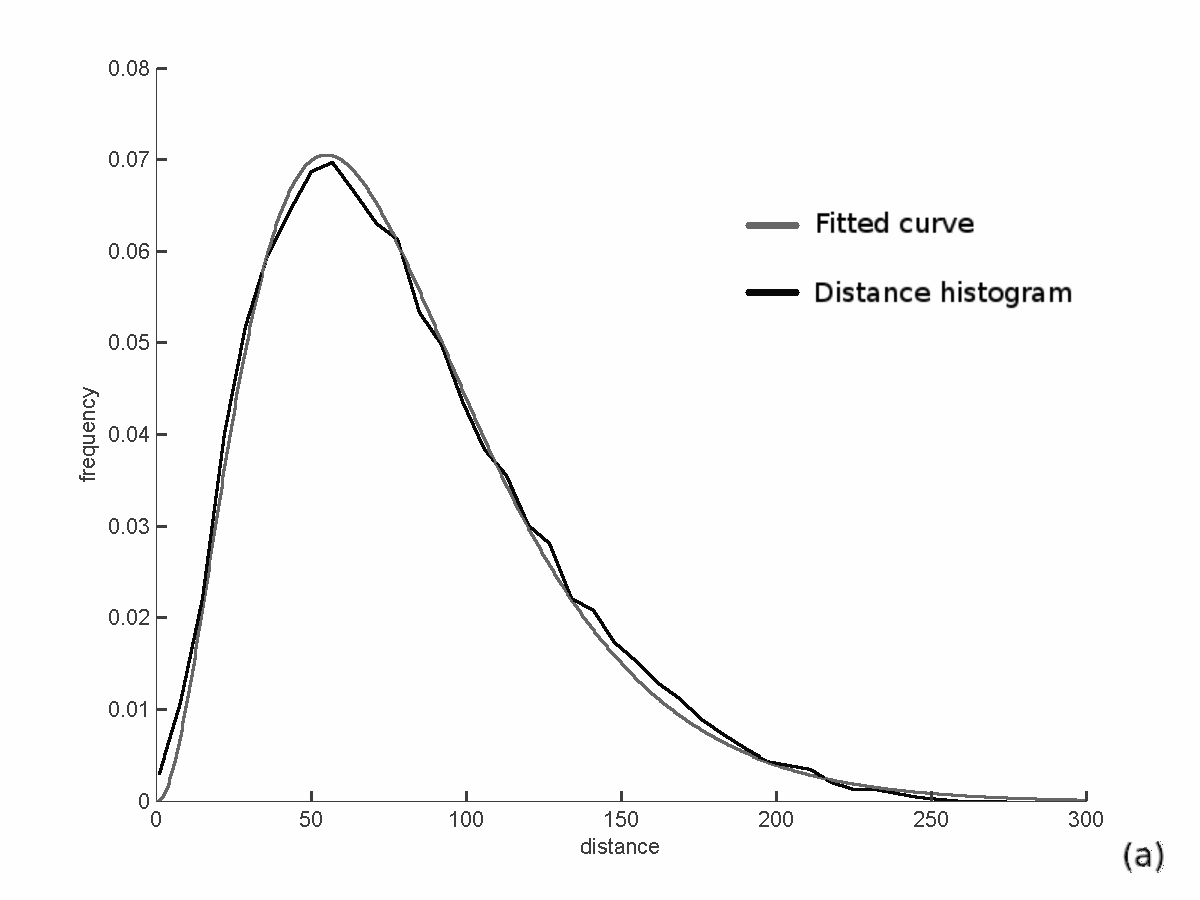,width=7cm}}

\end{minipage}
\begin{minipage}[h]{1 \linewidth}
  \centering
  \centerline{\epsfig{figure=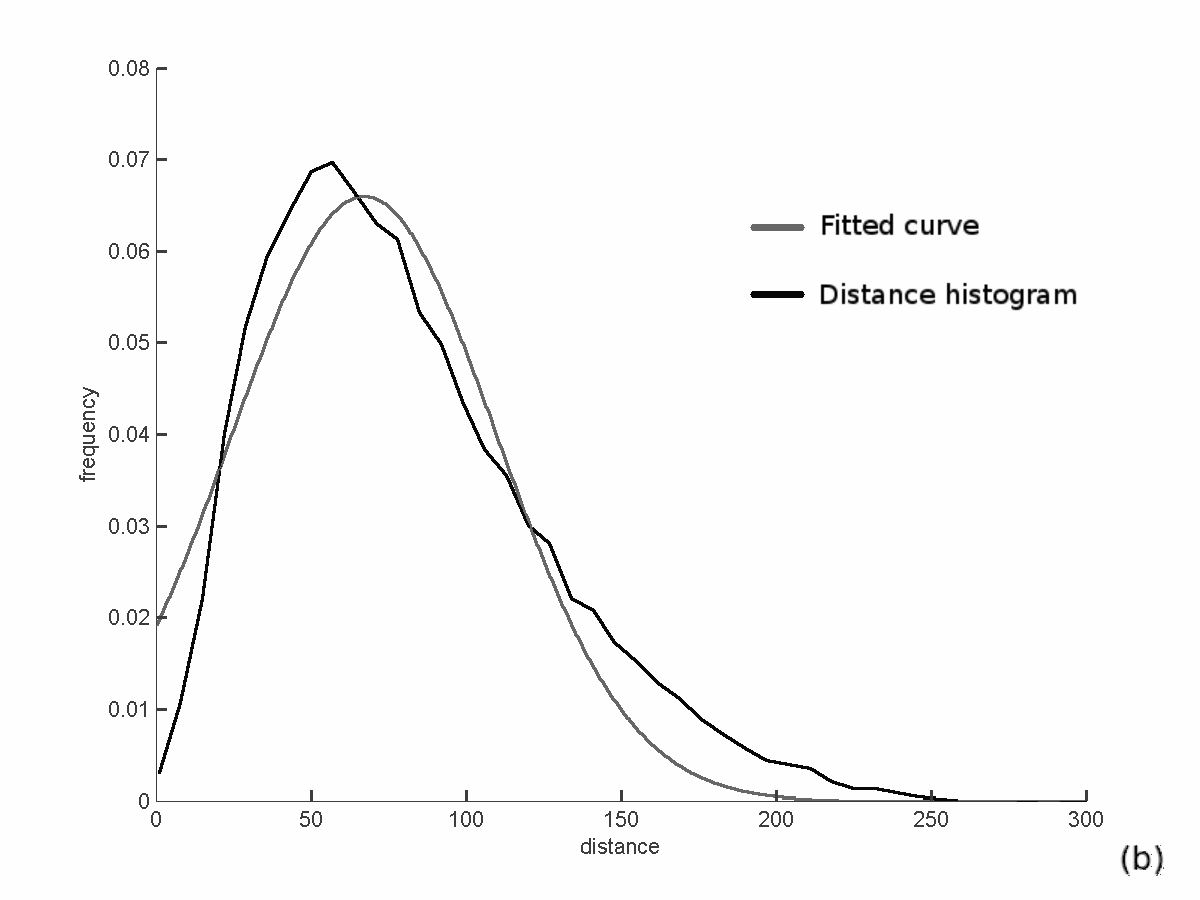,width=7cm}}

\end{minipage}

\begin{minipage}[h]{1 \linewidth}
  \centering
  \centerline{\epsfig{figure=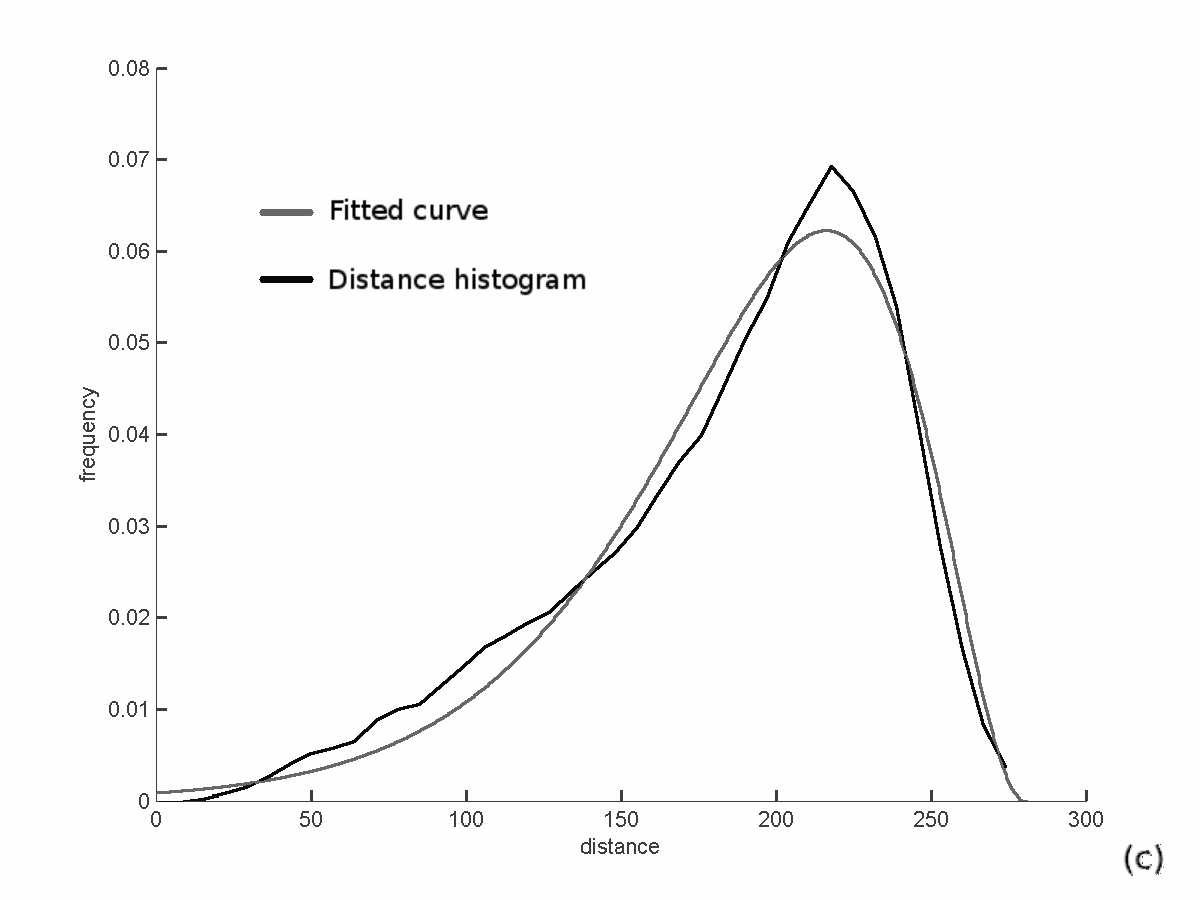,width=7cm}}

\end{minipage}
\begin{minipage}[h]{1 \linewidth}
  \centering
  \centerline{\epsfig{figure=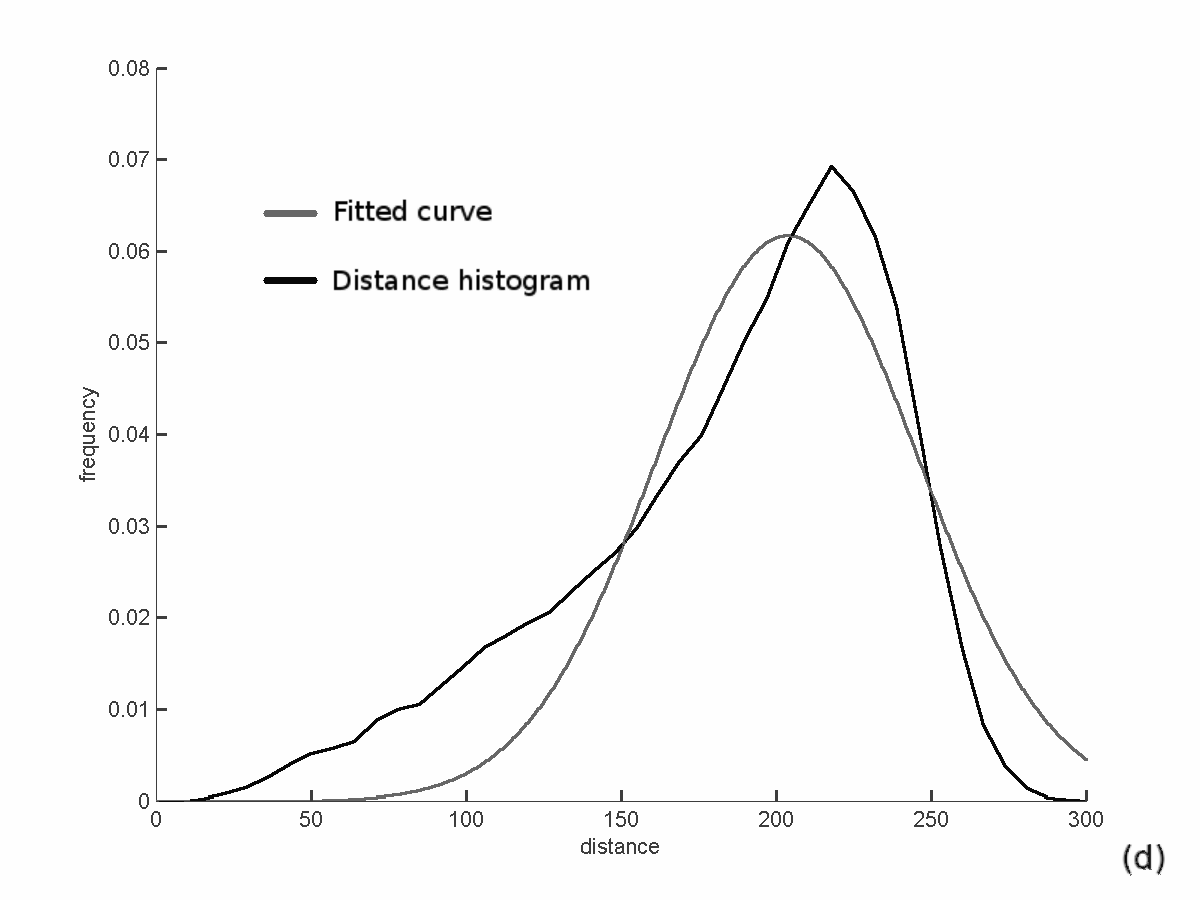,width=7cm}}

\end{minipage}

\caption{The correct (a and b) and incorrect (c and d) histograms of matching distances, fitted by a Chi (a and c) and a Normal (c and d) distribution.}
\end{figure}

\subsection{Statistical Modelling}

The histograms curves can be fitted by a distribution. On SIFT case, as we can see in Figure 2, the distance histograms are not symmetric. A frequently used probability distribution function, like the normal distribution, does not fit well the data (Figures 2b and 2d).  A non-symmetric distribution, would be better, but which one ? We have tested several distributions of similar "shape", among which Chi, Chi-square, Weibull and log-normal, and ended up selecting Chi (Figures 2a and 2c). Not only the Chi-distribution has the best overall fitting, but also it has the most satisfactory generative explanation for SIFT, since a Chi-distribution may be considered as the square-root of the sum of squared independent normals. Since SIFT has a L2-normalization, the Chi-distribution might easily arise. On our experiments the Chi function has the best fit among others. The fit was computed by the non-linear minimum square method. 

\subsection{Bayesian Search}

Combining the Bayesian decision theory \cite{BT} and the statistical model for the match distances we can retrieve a near-duplicate image with high accuracy and few feature vectors from the query.

Let $P(X)$ be the prior probability that a match is correct (i.e., the probability that the nearest neighbor from a query feature will match a correct image), which can be obtained during the training phase.

So $P(\overline{X})$ is the prior probability that a match is incorrect and $P(X) = 1-P(\overline{X})$.

$P(D|X)$ and $P(D|\overline{X})$ are the probabilities that a distance is D given the match is correct and incorrect, respectively.

Using the statistical model trained after the matches histograms, those probabilities can be computed analytically.

The likelihood that a single match $i$ is correct is:
$$
L_i = \frac{P(D_i|X)+E}{P(D_i|\overline{X})+E}
$$
 
\noindent $E$ is a small amount to avoid division by zero. 
The probability that an image $j$ is correct after $N$ matches from random samples is:

$$
	P_j(X|D_1\cap D_2 \ldots \cap D_N) = \frac{\prod_{i=1}^N{L_i}\times P(X)}{\prod_{i=1}^N{L_i}\times P(X) + P(\overline{X})}
$$		    
			
We can then define an upper threshold for $P_j(X|D_1\cap D_2 \ldots \cap D_N)$ for which no more samples are taken and the image $j$ is retrieved with desirable accuracy.

\section{RESULTS}

We have used the Bayesian search described above using 110,000 images for the noise database, all of them from the Yahoo database. The query images were generated by transformations of 225 target images. Each target image generated one query image. Half of the target images were from a cell phone (resolution X) and the other half from a digital camera (resolution Y). The transformations used were: crop, shear, rotation, re-scales, Gaussian noise and dithering. Figure 3 shows some examples of query image generation. The Euclidian distance was used for matching. Half of the query images were taken for test the Bayesian search, while the other half was used to generate the statistical model.

As we can see in Table 1, the Bayesian search had high accuracy using, on average, less than a dozen (from hundreds) feature vectors per query.
Even if the normal distribution needed less samples for a determined probability threshold to retrieve an image, its accuracy is notably worst than using chi distribution.

The lower number of samples while using the normal distribution fit is due to the low intersection between the correct and incorrect curves, but the tail of the histogram isn't well represented on this way and an amount of false positive is not detected.

\begin{figure}[htb]

\begin{minipage}[b]{1.0\linewidth}
  \centering
  \centerline{\epsfig{figure=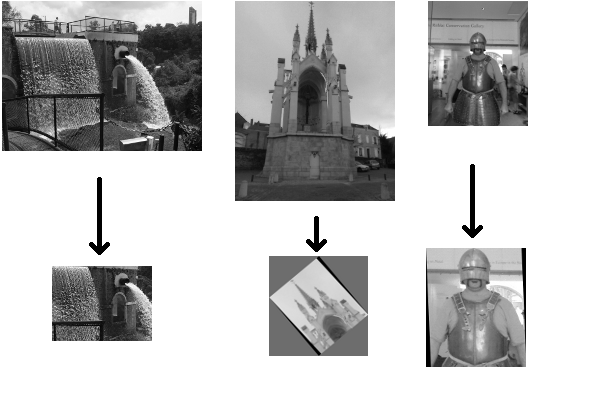,width=8cm}}

\end{minipage}

\caption{Examples of query images (below) generated by transforming its original images (above).}
\end{figure}

\begin{table}
\caption {accuracy and mean number of feature vector samples on Bayesian search with three probability thresholds}

\begin{tabular}{|c|c|c|c|c|}
\hline
%% row 1

&Threshold
&90.00\%
&99.00\%
&99.90\%
\\
\hline
%% row 2

Normal fit
&Mean $N$
&2.5
&3.9
&6.2
\\
\hline
%% row 3

Normal fit
&Accuracy
&85.5\%
&90.8\%
&97.3\%
\\
\hline
%% row 4

Chi fit
&Mean $N$
&3.8
&6.8
&10.2
\\
\hline
%% row 5

Chi
&Accuracy
&93.4\%
&98.7\%
&99.3\%
\\
\hline
\end{tabular}

\end{table}

\section{DISCUSSION}

It is striking (Figure 2) how tight the Chi distribution fits to the distances histograms. We believe that this might be an important observation, in general, for the use of SIFT features, beyond nearest-duplicate detection. Our results demonstrate that the benefits of the well-adjusted model can be quite dramatic, as the much better adjustment between expected (header) and observed (3rd line for Gaussian, 5th line for Chi) accuracies on Table 1 show.

\bibliographystyle{IEEEbib}
\bibliography{icip2011}

\end{document}